# Comparison of derivative-free and gradient-based minimization for multi-objective compositional design of shape memory alloys

S. Josyula, Y. Noiman, E. J. Payton[*], T. Giovannelli

*Department of Mechanical and Materials Engineering, University of Cincinnati, 2901 Woodside Drive, Cincinnati, OH 45221-0072*

**Abstract**

Designing shape memory alloys (SMAs) that meet performance targets while remaining affordable and sustainable is a complex challenge. In this work, we focus on optimizing SMA compositions to achieve a desired martensitic start temperature (Ms) while minimizing cost. To do this, we use machine learning models as surrogate predictors and apply numerical optimization methods to search for suitable alloy combinations. We trained two types of machine learning models, a tree-based ensemble and a neural network, using a dataset of experimentally characterized alloys and physics-informed features. The tree-based model was used with a derivative-free optimizer (COBYLA), while the neural network, which provides gradient information, was paired with a gradient-based optimizer (TRUST-CONSTR). Our results show that while both models predict Ms with similar accuracy, the optimizer paired with the neural network finds better solutions more consistently. COBYLA often converged to suboptimal results, especially when the starting guess was far from the target. The TRUST-CONSTR method showed more stable behavior and was better at reaching alloy compositions that met both objectives. This study demonstrates a practical approach to exploring new SMA compositions by combining physics-informed data, machine learning models, and optimization algorithms. Although the scale of our dataset is smaller than simulation-based efforts, the use of experimental data improves the reliability of the predictions. The approach can be extended to other materials where design trade-offs must be made with limited data.

---
[*] *Corresponding author. E-mail: paytonej@ucmail.uc.edu; Tel: +1-513-556-0260; Fax: +1-513-556-3390*

## 1. Introduction

The applications of NiTi alloys in automotive, aerospace, manufacturing at high temperatures can be limited by the relatively low transformation temperatures (TTs), under 100 °C. Irreversible changes in the material's microstructure caused by the temperature-driven martensitic transformation led to inefficiencies that limit their effectiveness as actuators or in thermal energy storage systems [1], [2], [3]. Cu-based shape memory alloys are appealing because of their low cost and high transformation temperatures; however, their functional stability and reliability are often limited. In contrast, while Fe-based shape memory alloys can exhibit excellent shape memory properties, they typically undergo a non-thermoelastic transformation, resulting in a larger thermal hysteresis [4]. This challenge of expanding the transformation temperature range of SMAs has led to a substantial body of research toward the development of High-Temperature Shape Memory Alloys (HTSMAs). Much work has focused on the addition of elements such as Cu, Au, Pd, Pt, V, Cr, Hf, and Zr to binary NiTi to improve the thermal hysteresis [5], [4].

However, the vast compositional space available due to the multitude of alloying elements presents a significant challenge in terms of experimentation. With hundreds of possible combinations of elements and varying concentrations, traditional experimental methods, while valuable, become impractical for exploring such a large parameter space.

This is where computational optimization methods offer critical advantages. By leveraging computational tools, researchers can simulate the thermodynamic behavior and phase transformations of numerous alloy compositions without the need for physical synthesis, drastically reducing time and costs. Global optimization in the context of HTSMAs refers to systematically searching the vast compositional space to identify optimal alloy compositions that are needed. Interactions between multiple elements often lead to non-linear relationships and affect the material properties; therefore, methods that can handle this level of complexity are utilized.

While artificial neural networks (or simply neural networks) are commonly used in machine learning for tasks like regression and classification, their application as surrogate models for optimization has been limited. In [6], the authors assessed the accuracy of neural networks as surrogate models for approximating the objective functions of optimization problems in CUTEst, a well-known library of nonlinear optimization test problems. The paper explores the performance of different activation functions, finding ReLU and SiLU to be the most effective. It also compares neural networks with traditional interpolation/regression models, revealing that neural networks can provide competitive approximations for function values and gradients. The research identifies some limitations of neural network surrogate models. First, more complex architectures require a large number of training points to achieve satisfactory accuracy, and this demand grows in higher-dimensional problems (i.e., problems with many optimization variables, in the order of hundreds). Second, when training points are unevenly distributed, the quality of approximations for function values and gradients significantly declines. In our paper, these limitations are not critical, as the problems are not large-scale, with fewer than 40 optimization variables. Additionally, we will impose constraints to ensure that the combinations of variables explored by the optimization algorithms remain sufficiently close to the training dataset.

Using neural network surrogate models to approximate computationally expensive black-box objective functions (i.e., those without an analytical expression and requiring costly simulations or experiments for evaluation) in engineering problems is not new. In [7], convolutional neural networks were employed to approximate finite-element simulations and predict the mechanical properties of periodic structures. In [8], the authors used neural network surrogates to approximate computationally expensive objective functions in geological CO2 storage operations, with the resulting multi-objective problem being solved using an evolutionary algorithm. In [9], the authors addressed a simulation-based optimization problem in healthcare by using a neural network to predict the output of a simulation model. This enabled the optimization process to be accelerated using a derivative-based algorithm rather than a derivative-free one. The neural network surrogate was trained on a dataset generated by running the simulation model prior to the optimization.

However, research exploring neural network surrogate models for materials applications is limited. In [10], a neural network was used to predict the behavior of carbon nanotubes under geometric nonlinearities. In [11], the authors proposed both polynomial and neural network surrogates to predict the rolling load in the cold rolling of flat metals. In [12], a neural network model was used to predict the behavior of shape memory alloys, particularly their superelastic properties, and explored their application in structural vibration control. In [13], the authors provide a comprehensive review of the use of neural networks in modeling shape memory alloys, highlighting their importance in various fields such as medical devices, robotics, aerospace, and civil engineering.

None of the research mentioned above focuses on alloy discovery, which is the gap that our paper aims to bridge.

*1.1 Statement of the optimization problem*

The optimization problem we propose aims to discover optimal alloy compositions to achieve the desired martensite start temperature while minimizing cost, resulting in two objective functions to minimize. We will use a machine learning model to predict the martensite start temperature for a given alloy composition. We denote each alloy composition as a vector $x \in \mathbb{R}^{39}$, where 39 represents the number of elements in the alloy, and each component $x_i$ corresponds to the percentage of the $i$-th element in the composition, with $i \in \{1, ..., 39\}$, such that $\sum_{i=1}^{39} x_i = 100$. We denote as $y(x) = (y_1(x), ..., y_7(x))' \in \mathbb{R}^7$ the column vector that associates seven physics-based features with each alloy composition $x$, where the prime symbol denotes the transpose operator. Such features are enthalpy of mixing ($\Delta H_{mix}$), lattice constant ($a$), valence electron concentration ($vec$), atomic radius (mean $\bar{r}$, variation coefficient $\delta r$), and electronegativity (mean $\bar{\chi}$, variation coefficient $\delta \chi$). Note that each physics-based feature can be seen as a scalar function of $x$ and can be computed as follows:

($\Delta H_{mix}$): $y_1(x) = \sum_{i \in \{1,...,39\}} \sum_{j \in \{1,...,39\}, j>i} 4 \frac{x_i}{100} \frac{x_j}{100} H_{mix}^{ij}$, with $H_{mix}^{ij} \sim \frac{\bar{H}_0}{4}$, where $\bar{H}_0$ are two limiting heats of solution.

($a$): $y_2(x) = \sum_{i=1}^{39} x_i (\frac{n_i M}{\rho N_A})^{1/3 \, 1/3}$, where $n$ is the number of atoms per unit cell, $M$ is the molar mass, $\rho$ is the density, and $N_A$ is Avogadro's number.

($vec$): $y_3(x) = \frac{\sum_{i=1}^{39} \frac{x_i}{100} e_v^i}{\sum_{i=1}^{39} \frac{x_i}{100} Z_i}$, where $e_v^i$ is the atomic number and $Z_i$ is the valence electron.

($\bar{r}$): $y_4(x) = \sum_{i=1}^{39} \frac{x_i}{100} r_i$, where $r_i$ is the atomic radius in angstroms (Å)

($\delta r$): $y_5(x) = \sqrt{\sum_{i=1}^{39} \frac{x_i}{100} (1 - \frac{r_i}{\bar{r}})^2}$.

($\bar{\chi}$): $y_6(x) = \sum_{i=1}^{39} \frac{x_i}{100} \chi_i$, where $\chi_i$ is the electronegativity of pure elements.

($\delta \chi$): $y_7(x) = \sqrt{\sum_{i=1}^{39} \frac{x_i}{100} (1 - \frac{\chi_i}{\bar{\chi}})^2}$.

The machine learning model takes the feature vector $y(x)$ as input. The model's response function evaluated at this input is denoted as $z(y(x))$. Given a target martensite temperature $T_s$, the first objective computes the relative squared error between $T_s$ and the temperature predicted by the machine learning model for the given alloy composition, denoted as $\widehat{T}_s(x) = z(y(x))$. Given the cost of element $i$, $C_i$, in \$/kg and the atomic mass $m_i$ in kg/mol for each element $i$, the second objective provides an estimate of the relative cost of the alloy. The analytical expressions for both objective functions, denoted as $f_1$ and $f_2$, are as follows:

$$f_1(x) = \left(\frac{T_s - \widehat{T}_s(x)}{T_s}\right)^2 \text{ and } f_2(x) = \sum_{i=1}^{39} C_i \frac{x_i m_i}{(\sum_{l=1}^{39} x_l m_l)}.$$

Since the optimization problem we aim to solve is bi-objective, we will apply the well-known weighted-sum method, which is a classical approach for multi-objective optimization that consists of weighting the objective functions into a single objective [14], [15]. We also need to ensure that the optimization variables $x_i$, with $i \in \{1, \ldots, 39\}$, represent percentages, which can be accomplished by using a functional equality constraint:

$$g_1(x) = \sum_{i=1}^{39} x_i - 100 = 0.$$

Given non-negative weights $\lambda_1$ and $\lambda_2$, such that $\lambda_1 + \lambda_2 = 1$, the resulting optimization problem to solve is

$$\text{Min } f(x) = \lambda_1 f_1(x) + \lambda_2 f_2(x)$$

$$\text{s.t. } g_1(x) = 0.$$

*1.2 Description of the surrogate-based optimization methodology*

The choice of optimization method to solve the problem in the previous section varies significantly depending on the machine learning model employed. In this paper, we will predict the martensite temperature using two machine learning models: a tree-based ensemble model and an artificial neural network. When using the tree-based ensemble model, which is defined by a non-differentiable, piecewise-constant response function, we will solve the optimization problem using a derivative-free optimization algorithm [16], [17], [18]. In contrast, when using the artificial neural network, which has a differentiable response function, we will use a derivative-based optimization algorithm from nonlinear optimization [19].

Specifically, when using the artificial neural network, the gradient for the first objective function (i.e., the 39-dimensional vector of all partial derivatives) can be computed by applying the chain rule, resulting in:

$$\nabla f_1(x) = -2\left(T_s - \widehat{T}_s(x)\right) \nabla_x y(x) \nabla_y z(y(x)).$$

In this equation, $\nabla_x y(x) = (\nabla_x y_1(x), \ldots, \nabla_x y_7(x))$ is the 39 × 7 Jacobian matrix of $y(x)$, with each column representing the gradient of the corresponding scalar function $\nabla_x y_j(x)$, for $j \in \{1, \ldots, 7\}$, and $\nabla_y z(y(x))$ is the 7-dimensional gradient of the neural network response function with respect to the variables $y$, evaluated at $y(x)$. The gradient $\nabla_y z(y(x))$ can be efficiently computed via backpropagation, which is implemented using PyTorch's autograd library, an automatic differentiation tool. When using the tree-based ensemble model, the response function is non-differentiable and piecewise-constant, making the gradient of the first objective function either undefined (at points of non-differentiability) or uninformative (identically zero elsewhere).

Note that the gradient of the second objective function can always be computed analytically, independent of the choice of machine learning model.

To mitigate potential inaccuracies in predicting martensite temperature for alloy compositions far from the training data, we impose a constraint ensuring that the optimization algorithm explores compositions sufficiently close to the training dataset. Let $D = \{(y(x^l)', T_s(x^l)) \mid l \in \{1, \ldots, N\}\}$ represent our dataset, where $N$ denotes the dataset size, and $l$ indexes a row in the dataset, corresponding to an alloy composition $x^l \in \mathbb{R}^{39}$. The following constraint ensures that the minimum distance between a given alloy composition and any alloy composition in the dataset in terms of $y$ remains below a threshold $\tau$, as follows:

$$g_2(x) = \min_{l \in \{1,\ldots,N\}} \| y(x) - y(x^\wedge l) \|_2 - \tau \leq 0,$$

where $\|\cdot\|_2$ denotes the $\ell$-2 norm of the given vector (i.e., the square root of the sum of the squared components).

Before running the optimization process, the tree-based ensemble model and the artificial neural network will be trained on a dataset containing alloy compositions with the seven physics-based features mentioned previously. Then, when using the tree-based ensemble model, we will solve the optimization problem using COBYLA [20], a derivative-free optimization algorithm that supports constraints through a penalization approach, treating them as soft (i.e., they may be violated). Since COBYLA is not designed to handle equality constraints, a projection operator onto the set $\{x \in R^{39} \mid g_1(x) = \sum_{i=1}^{39} x_i - 100 = 0\}$ is applied to ensure that the optimization variables $x_i$, with $i \in \{1, \ldots, 39\}$, represent percentages, leading to feasible alloy compositions. When employing a neural network model, we will use TRUST-CONSTR [21], a derivative-based trust-region algorithm that enforces constraints at each iteration, treating them as hard constraints. TRUST-CONSTR supports equality constraints, allowing the constraint $g_1(x) = 0$ to be enforced without the need for projections.

When using either model, we normalize the objective functions in $f(x)$ dividing $f_1(x)$ and $f_2(x)$ by their respective values at the initial point of each optimization process, denoted as $x_0$. This standard approach in the multi-objective optimization literature prevents one objective from becoming dominant due to differences in scaling [22]. Therefore, the actual optimization problem to solve is

$$\text{Min } f(x) = \lambda_1 \frac{f_1(x)}{f_1(x_0)} + \lambda_2 \frac{f_2(x)}{f_2(x_0)}$$

$$s.t. \; g_1(x) = 0 \text{ and } g_2(x) \leq 0.$$

## 2. Materials & methods

The dataset is sourced from NASA created by Benafan et al [23], a physics-informed dataset by Thiercelin et al [24], and our own unique dataset consisting of ternary, quaternary and quinary SMAs. The dataset size is $N = 6{,}708$.

There are two challenges in this dataset. The first challenge is that different alloys can have the same martensite temperature, i.e., there exist $l_1$ and $l_2$ in $\{1, \ldots, N\}$, with $l_1 \neq l_2$, such that $y(x^{l_1}) \neq y(x^{l_2})$ and $T_s(x^{l_1}) = T_s(x^{l_2})$. This is an inherent issue in the dataset, and there's little we can do about it. It poses a challenge when validating the accuracy of machine learning models because it leads to multiple local minimizers for the first objective function in our problem. Specifically, when setting $\lambda_1 = 1$ and $\lambda_2 = 0$ (i.e., minimizing only the first objective), if multiple alloys have the same martensite temperature, it becomes unclear which alloy the optimization will return for a given target temperature. We will analyze this issue in detail when minimizing the first objective function in our optimization problem.

The second challenge is the presence of duplicated alloys associated with different martensite temperatures, i.e., there exist $l_1$ and $l_2$ in $\{1, ..., N\}$, with $l_1 \neq l_2$, such that $y(x^{l_1}) = y(x^{l_2})$ and $T_s(x^{l_1}) \neq T_s(x^{l_2})$. This is problematic because the predicted temperatures may be inaccurate due to conflicting data. The presence of duplicates arises because each alloy in the dataset was tested under different experimental conditions. Unlike the first challenge, the second challenge can be easily addressed by removing the duplicates and taking the median of the martensite temperatures associated with the duplicated alloys. However, this reduces the dataset size from 6,708 to 1,202. It is important to note that, with fewer data points, the predicted temperatures may still be inaccurate. It remains unclear which dataset leads to higher accuracy (whether the full dataset of 6,708 entries or the reduced one with 1,202), so we test both datasets.

When using the original dataset, we split the data into 80% for training and 20% for testing. We selected an extremely randomized trees model as our tree-based ensemble method, which builds multiple decision trees and makes predictions by averaging their outputs. We also tested a random forest regressor and a gradient boosting regressor. The extremely randomized trees model yielded results on the testing data that were comparable to those of the random forest regressor and outperformed the gradient boosting regressor. The extremely randomized trees model was implemented using the ExtraTreesRegressor from the Scikit-Learn library, with 50 estimators (trees) and a maximum depth of 100 for each tree. For the neural network model, we implemented a fully connected architecture with three layers in the PyTorch library. The input layer consists of 7 features, which are passed through two hidden layers: the first hidden layer has 64 neurons, and the second hidden layer has 32 neurons. Each hidden layer uses the ReLU activation function. The output layer consists of a single neuron to predict the target martensite temperature. The network was trained using the Adam optimizer with 2000 epochs, a batch size of 64, a learning rate of 0.01, and a weight decay of 0.01.

When using the dataset without duplicates, we split the data into 70% for training and 30% for testing. The lower training percentage than the typical 80:20 split helps reduce overfitting due to the smaller dataset size. We again selected an extremely randomized trees model as our tree-based ensemble model, using 40 estimators (trees), a maximum depth of 15 for each tree, and enabling bootstrapping, which led to better results on the testing data. For the neural network model, we implemented a fully connected architecture with four layers in the PyTorch library. The input layer consists of 7 features, which are passed through three hidden layers: the first hidden layer has 128 neurons, the second has 64 neurons, and the third has 32 neurons. Each hidden layer uses the ReLU activation function. Additionally, a dropout layer with a dropout rate of 0.1 is applied after each hidden layer to help prevent overfitting. The output layer consists of a single neuron. The network was trained using the Adam optimizer with 1000 epochs, a batch size of 128, a learning rate of 0.01, and a weight decay of 0.01.

We ran COBYLA and TRUST-CONSTR with default parameters. The threshold for the $g_2(x) \leq 0$ constraint was set to 0.4.

Throughout the numerical experiments section, we will apply principal component analysis (PCA) [25] to generate 2D plots that visualize the seven physics-based features of the training dataset, along with those corresponding to the sequence of points generated by the optimization algorithms, including both the initial and final points of the optimization process. PCA is a dimensionality reduction technique that computes linear combinations of the dataset's feature values to preserve maximum variability. In the 2D plots, the x-axis represents the first principal component (PC1), while the y-axis represents the second principal component (PC2). In all PCA plots presented in this paper, the first two principal components capture approximately 60% of the total variability in the dataset.

To optimize for cost function, the distribution in our dataset and trends and potential areas for optimization in alloy selection was noted in Figure 1. To address the primary objective of optimizing the martensitic start temperature (Ms) while simultaneously minimizing the costs, we utilized the COBYLA (Constrained Optimization BY Linear Approximations) [26] algorithm. COBYLA is a derivative-free optimization method, ideal for problems where the objective function's derivatives are difficult or impractical to compute. This is particularly relevant in our case, where the objective function is complex and nonlinear, involving multiple alloying elements with intricate interdependencies. The optimization process aimed to identify optimal alloy compositions that balance the desired transformation temperature with minimal cost.

Along with COBYLA, we monitored the RHOBEG parameter (initial trust-region radius) to assess the convergence behavior of the optimization process. These plots provided a graphical representation of the optimization progress, illustrating how the algorithm explored the compositional space and converged to an optimal solution while tracking cost and Ms temperature. Monitoring RHOBEG helped us understand whether the optimization process was successfully converging toward a solution or getting stuck in local minima due to the complex and high-dimensional nature of the problem. The results of this are plotted in Figure 2 (a), which shows the relationship between the objective function value and RHOBEG values. The objective function value provides a measure of how well the current solution satisfies the problem's constraints. A lower objective function value, ideally on a log scale indicates a better solution. (b) represents the time taken for each optimization step against the RHOBEG values. The optimization process was initially slow and caused the program to crash multiple times and despite tracking the RHOBEG values and adjusting our optimization steps, the algorithm's performance fluctuated without converging to the optimal solution.

The complexity of optimizing for multiple objectives, such as cost, and martensite start temperature, combined with the sparse experimental data available, led to the emergence of local minima rather than a global solution. This limitation highlights the challenges in global optimization within the alloy design and the need for more robust and comprehensive optimization algorithms.

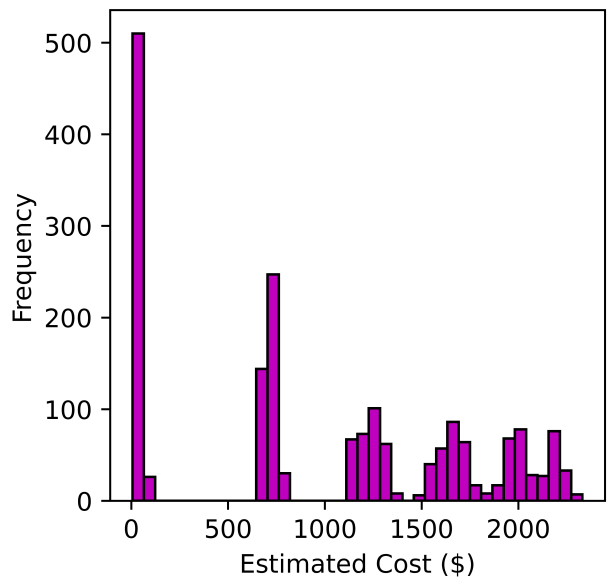

*Figure 1. Distribution of estimated alloy costs in the dataset*

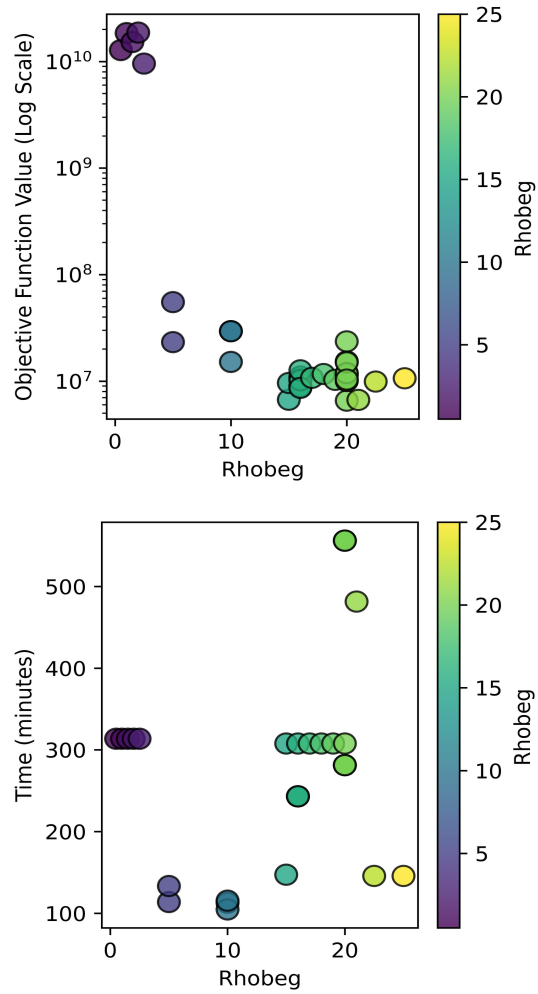

*Figure 2. (a) Log scale objective function value vs. RHOBEG (b) Time vs. RHOBEG for optimization process*

## 3. Results

*3.1 Assessing the accuracy of machine learning-based surrogate models*

We begin by assessing the accuracy of both the tree-based ensemble model and the neural network on the original dataset.

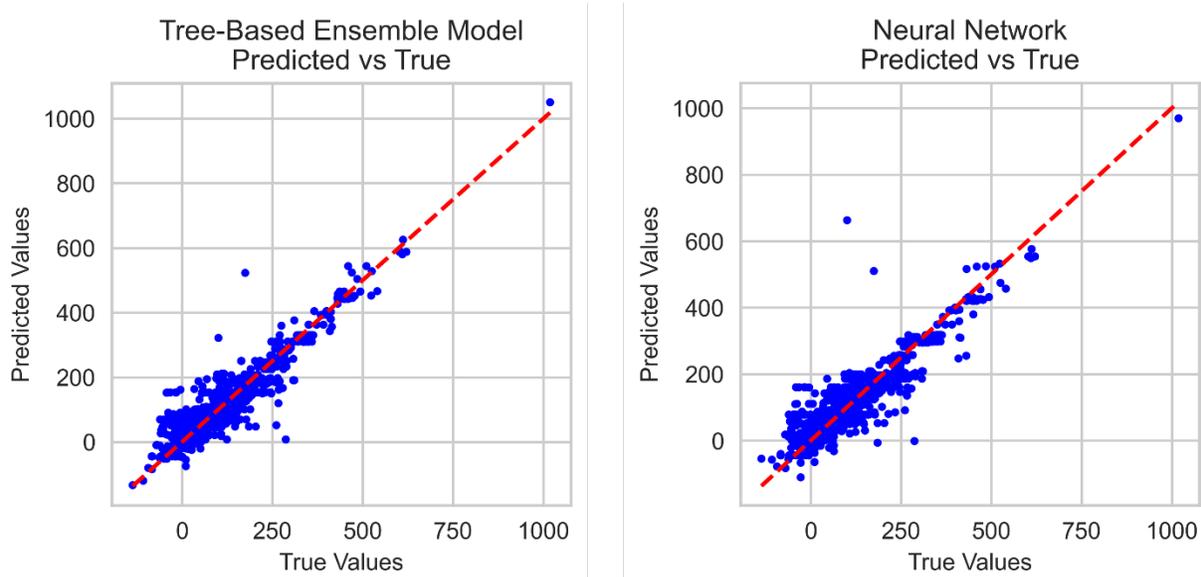

*Figure 3. Predicted martensite temperature vs. true martensite temperature for each data point in the original testing dataset, as predicted by the tree-based ensemble model (left) and the neural network (right).*

Figure 3 shows the comparison between the predicted martensite temperature and the true martensite temperature for each data point in the original dataset, as predicted by the tree-based ensemble model (left) and the neural network (right). The training and testing R-squared scores for the tree-based ensemble model are 92.91% and 87.28%, respectively, with a testing mean absolute error of 26.27. For the neural network, the training and testing R-squared scores are 88.95% and 83.43%, respectively, and the testing mean absolute error is 30.29. Although the neural network is slightly less accurate than the tree-based model, we do not expect this to be an issue during optimization, as the accuracy difference is marginal.

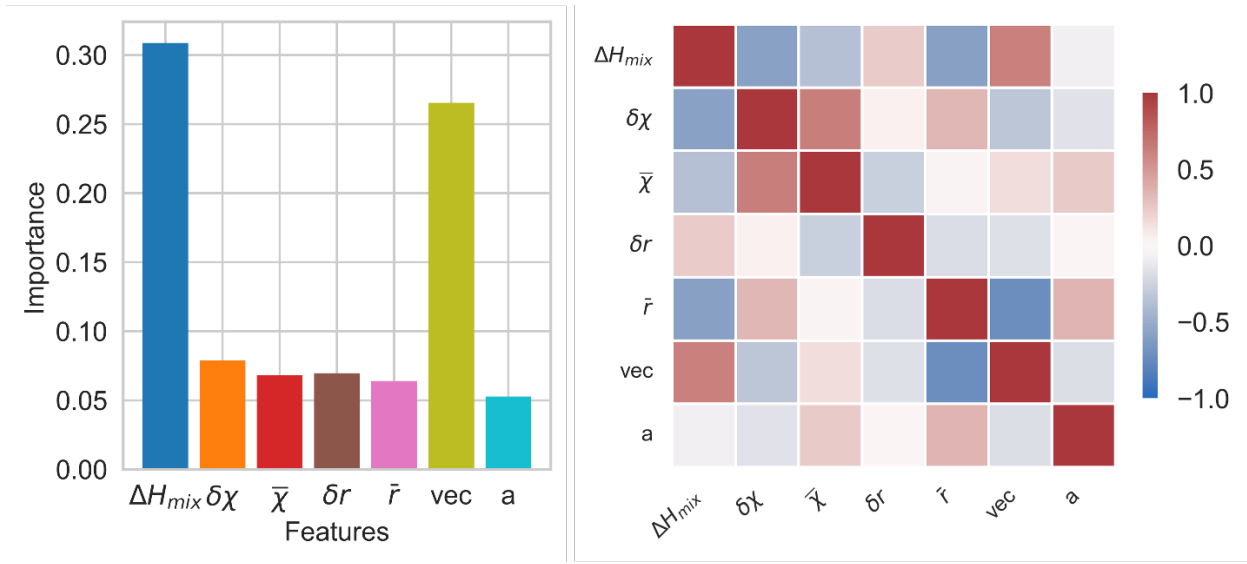

*Figure 4. Feature importance based on the original dataset (left), and pairwise correlations between physics-based features in the original dataset (right).*

Figure 4 displays the feature importance plot and the correlation among the seven physics-based features in the original dataset when using the tree-based ensemble model, indicating that $\Delta H_{mix}$ and $vec$ are the most important features for predicting the martensite temperature.

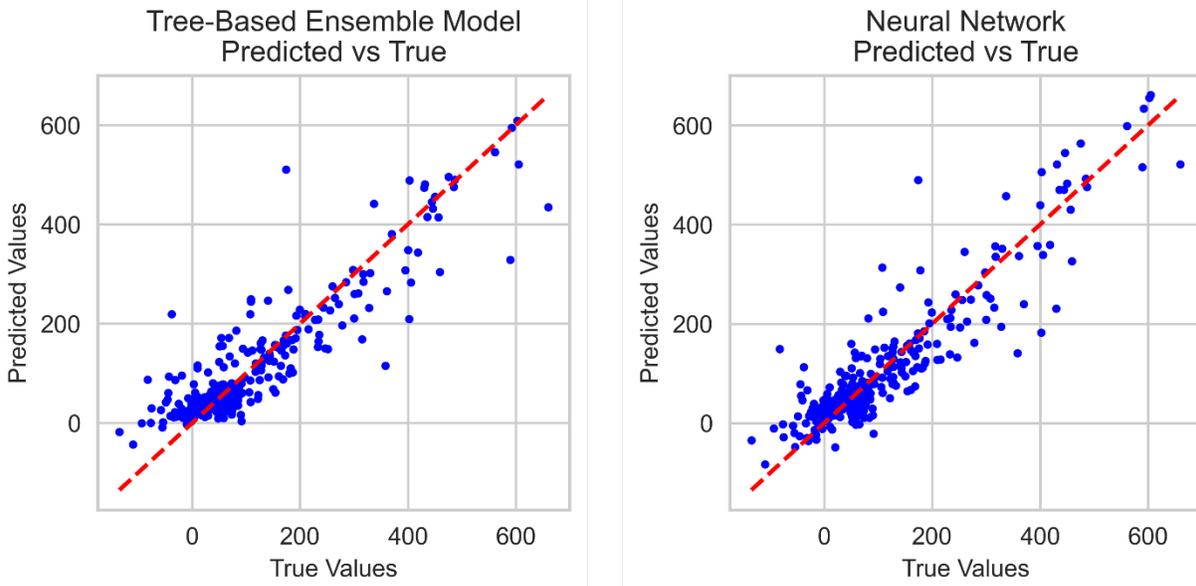

*Figure 5. Predicted martensite temperature vs. true martensite temperature for each data point in the testing dataset without duplicated alloys, as predicted by the tree-based ensemble model (left) and the neural network (right).*

Figure 5 shows the comparison between the predicted martensite temperature and the true martensite temperature for each data point in the dataset without duplicated alloys, as predicted by the tree-based ensemble model (left) and the neural network (right). The training and testing R-squared scores for the tree-based ensemble model are 96.82% and 81.30%, respectively, with a testing mean absolute error of 36.05. For the neural network, the training and testing R-squared scores are 92.17% and 82.67%, respectively, and the testing mean absolute error is 35.76. Therefore, on the dataset without duplicated alloys, both models show reduced accuracy compared to the original dataset due to the smaller number of data points. However, the decrease in accuracy is not significant for our purposes in the context of optimization. Therefore, for the remainder of this paper, we will use the dataset without duplicated alloys.

We do not include figures for the feature importance plot and the correlation among the seven physics-based features for the dataset without duplicated alloys, as they are very similar to those obtained for the original dataset.

*3.2 Achieving the desired temperature using machine learning-based surrogate models*

The goal of this subsection is to validate the tree-based ensemble model and the neural network in identifying the alloy composition that achieves the desired martensite temperature. To accomplish this, we will focus on minimizing only the first objective function by setting $\lambda_1 = 1$ and $\lambda_2 = 0$. As previously mentioned, one of the two challenges in our dataset is that different alloys can have the same martensite temperature, which leads to multiple local minimizers for the first objective function in our problem. Therefore, it becomes unclear which alloy the optimization will return for a given target martensite temperature. The result will depend on the proximity of the initial optimization point to the nearest local minimizer.

From the dataset, we arbitrarily selected the only alloy with a martensite temperature of 90.45 (see Figure 6). Although this is the only alloy in the dataset with this specific temperature, there are six other alloys with temperatures between 90 and 91, which could prevent the algorithm from converging to the alloy with a martensite temperature of 90.45.

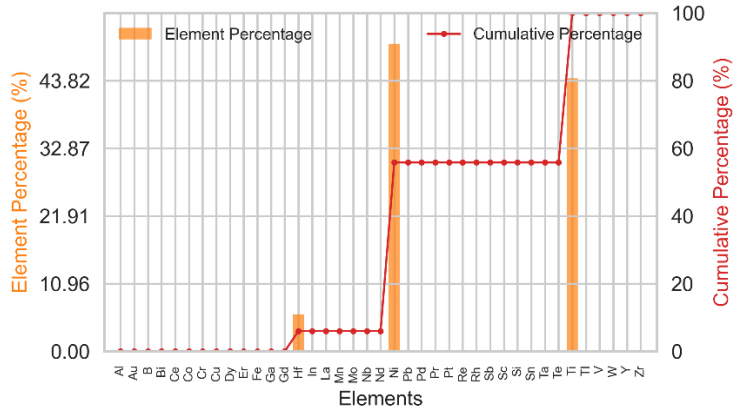

*Figure 6. Alloy composition with a martensite temperature of 90.45.*

As the initial point for the optimization, we used the alloy in Figure 6, applying random perturbations to the non-zero element percentages. These perturbations were sampled from a uniform distribution in the range $(-u, u)$, with $u > 0$. Ideally, by setting $T_s = 90.45$ in the first objective function, both optimization algorithms, COBYLA and TRUST-CONSTR, should converge to the alloy in Figure 6 or to another alloy with a martensite temperature of 90.45.

Figure 7 shows the results for COBYLA. Note that even when the perturbation is 0, the final alloy obtained by COBYLA is close to the initial alloy but with significant differences, as the percentage of Ni is lower, and the percentages of Al and Au are non-zero. When the perturbation is set to $u = 10$, the differences become more pronounced, with more elements having percentages greater than 0.

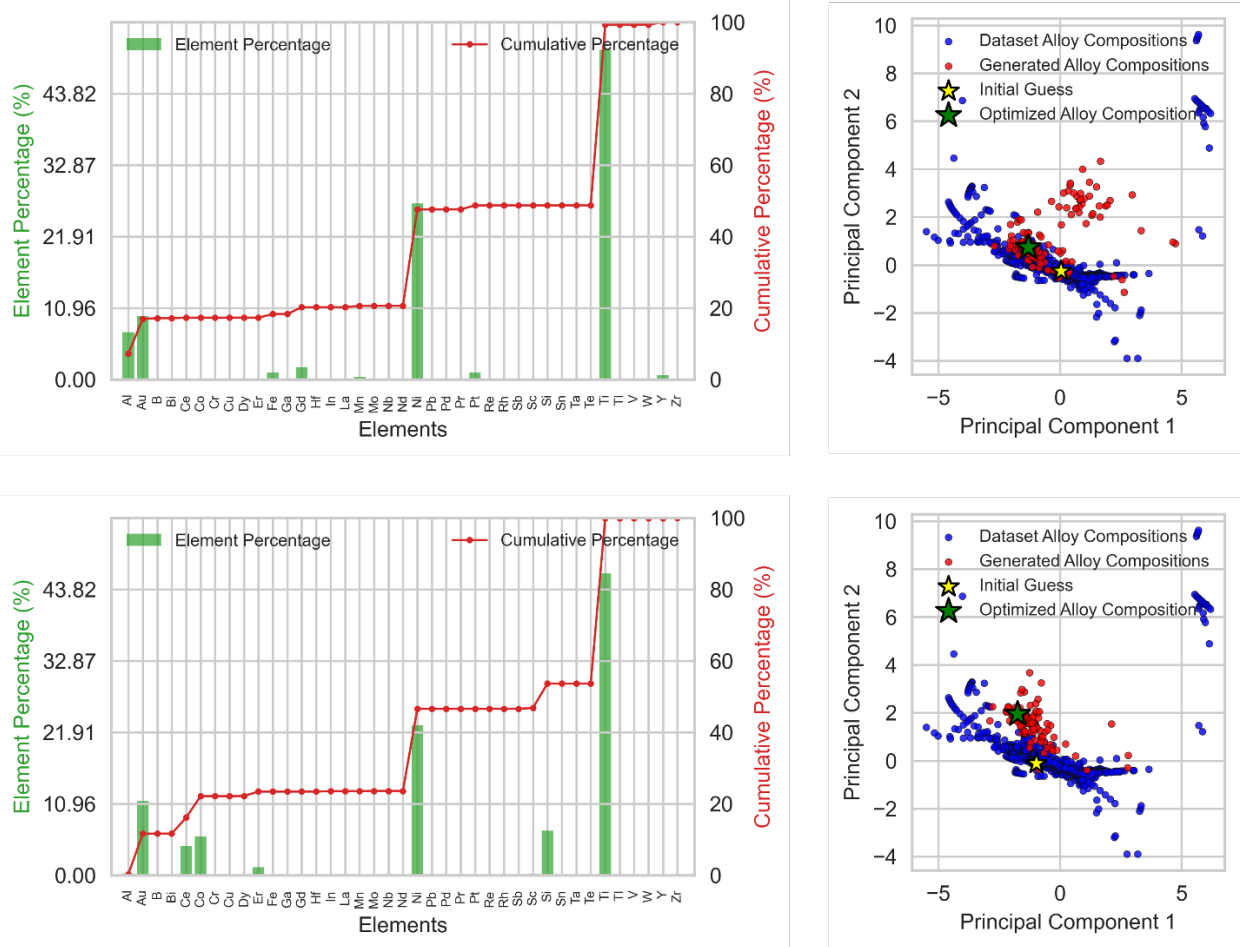

*Figure 7. Final alloy obtained using COBYLA with the initial alloy composition from Figure 6 (left), and visualization of the optimization process (right). The top row corresponds to $u = 0$ (no perturbation), and the bottom row to $u = 10$.*

Figure 8 show the results for TRUST-CONSTR. Note that when the perturbation is 0, the final alloy obtained by TRUST-CONSTR is approximately the same as the initial alloy (the percentage of Ti is lower), as expected. When the perturbation is set to $u = 10$, the final alloy is close to the initial one, but with some differences, as the percentages of Hf and Ni are lower than in the initial alloy. When the perturbation is set to $u = 20$, the final alloy is still quite close to the initial one, but with a higher percentage of Ga than in the initial alloy. Therefore, based on this experiment, using TRUST-CONSTR equipped with a neural network appears to be more reliable than using COBYLA with the tree-based ensemble model (although the predictions from the latter are slightly more accurate, as seen previously), as the derivative information contained in the gradient of the first objective function drives TRUST-CONSTR to the correct target alloy.

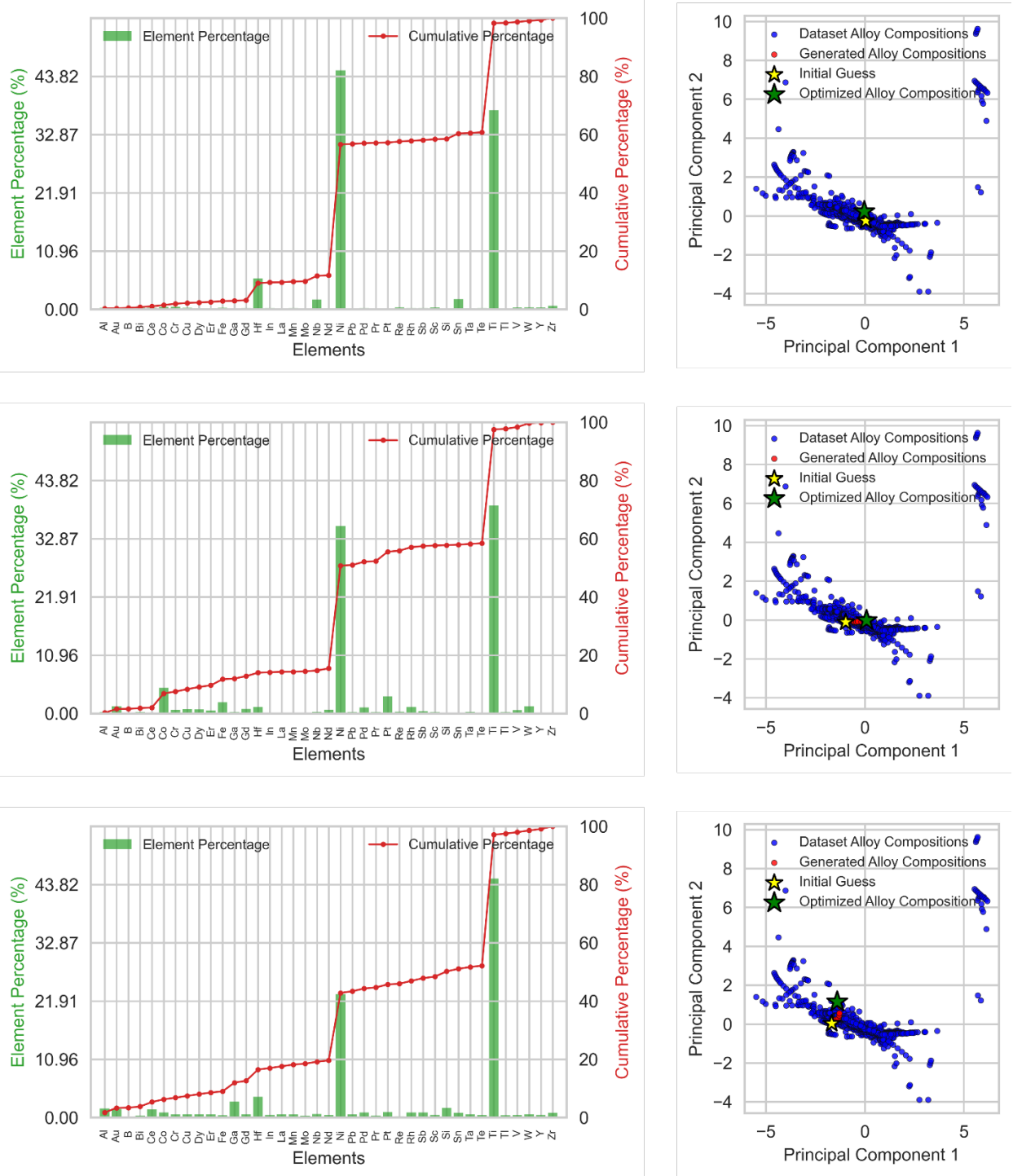

*Figure 8. Final alloy obtained using TRUST-CONSTR with the initial alloy composition from Figure 6 (left), and visualization of the optimization process (right). The top row corresponds to u=0 (no perturbation), the middle row to u=10, and the bottom to u=20.*

Now, let us repeat the experiment with a different initial alloy. This time, we selected one of the two alloys in the dataset with a martensite temperature of 100 (see Figure 9). There are two other alloys in the dataset with temperatures between 99.50 and 100.50.

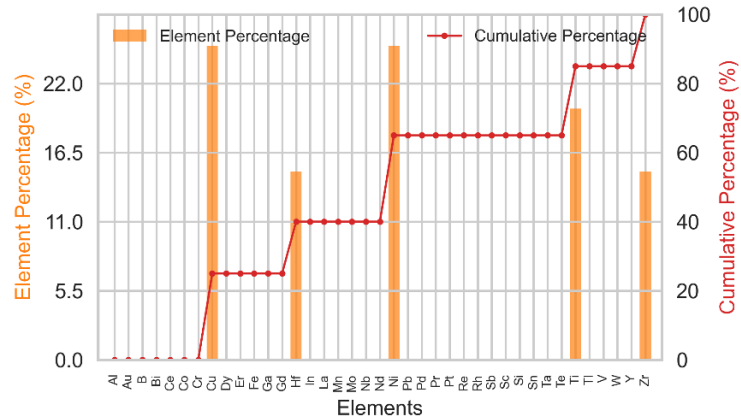

*Figure 9. Alloy composition with a martensite temperature of 100.*

As before, to obtain the initial point for the optimization, we applied random perturbations from a uniform distribution in the range $(-u, u)$ to the non-zero element percentages of the alloy in Figure 9.

Figure 10 below shows the results for COBYLA. This time, when the perturbation is 0, the final alloy obtained by COBYLA is very close to the initial alloy. When the perturbation is set to $u = 10$, the differences become much more pronounced.

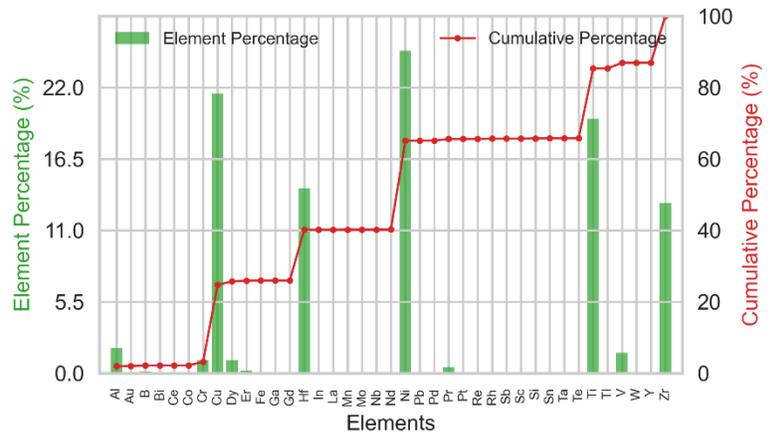 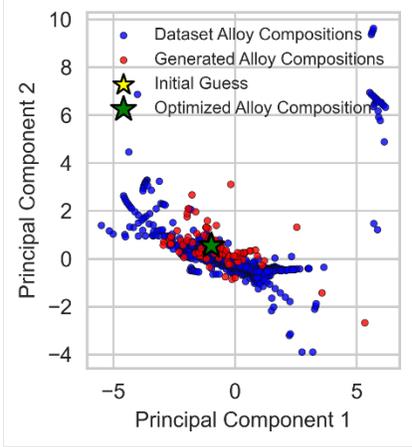

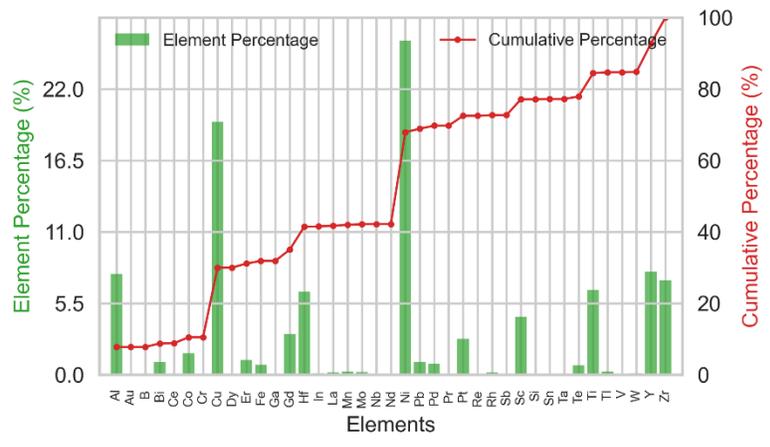 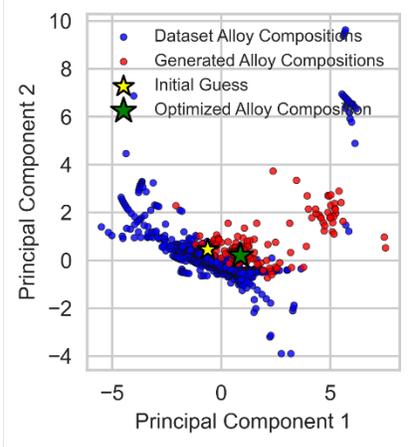

*Figure 10. Final alloy obtained using COBYLA with the initial alloy composition from Figure 9 (left), and visualization of the optimization process (right). The top row corresponds to $u = 0$ (no perturbation), and the bottom row to $u = 10$.*

Figure 11 below shows the results for TRUST-CONSTR. When the perturbation is 0, the final alloy obtained by TRUST-CONSTR is approximately the same as the initial alloy, similar to COBYLA. When the perturbation is set to $u = 10$, the final alloy differs from the initial one but remains closer to it than the alloy obtained by COBYLA. This confirms that the derivative information allows TRUST-CONSTR to converge to the target alloys more accurately than COBYLA.

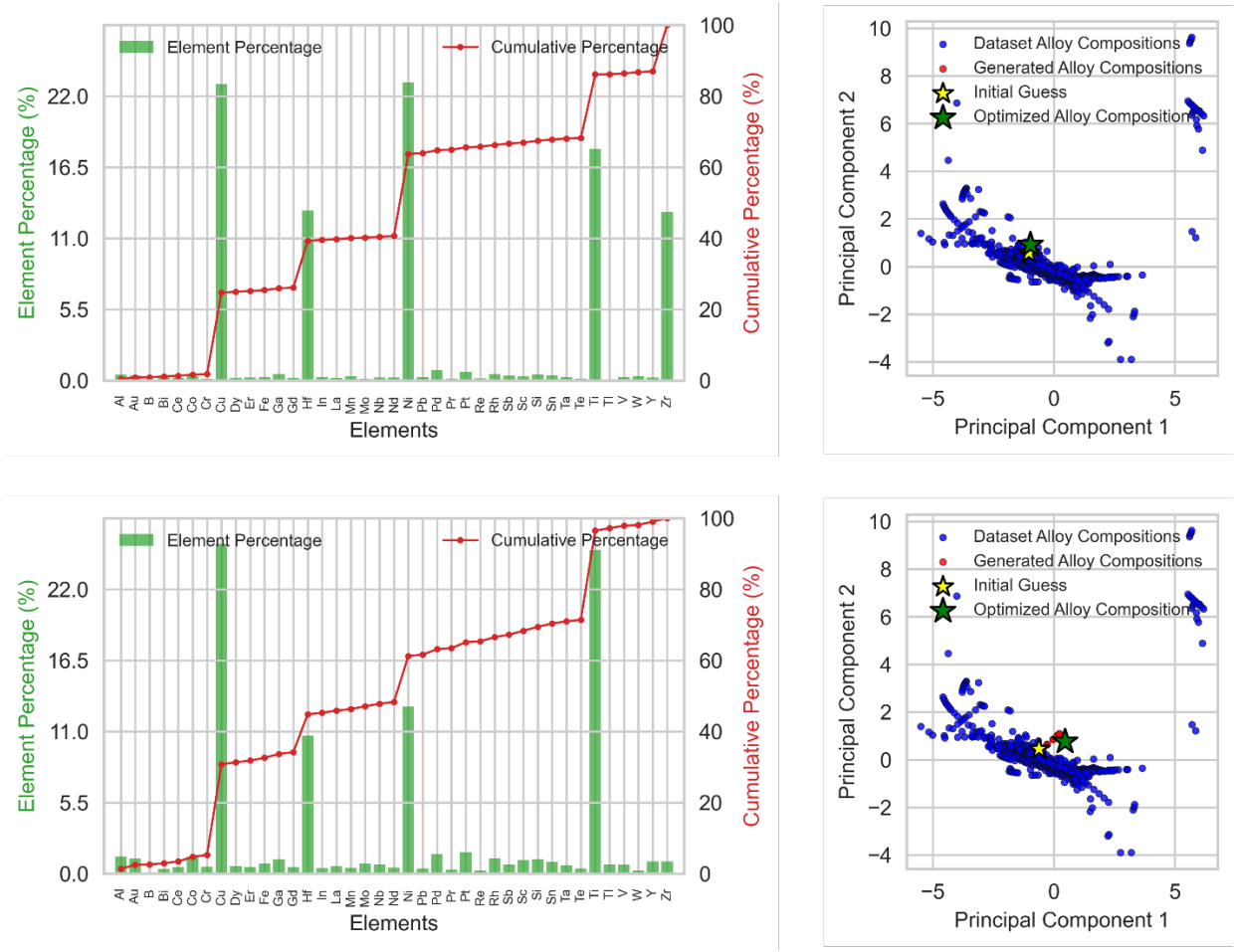

*Figure 11. Final alloy obtained using TRUST-CONSTR with the initial alloy composition from Figure 9 (left), and visualization of the optimization process (right). The top row corresponds to $u = 0$ (no perturbation), and the bottom row to $u = 10$.*

*3.3 Minimizing alloy cost*

We will now focus on minimizing only the second objective function by setting $\lambda_1 = 0$ and $\lambda_2 = 1$. Since the machine learning-based surrogate models are not required to evaluate the second objective, we will proceed without them. Given the presence of many local minimizers, we will apply each optimization algorithm using multiple initial alloys from the dataset. Specifically, for each algorithm, we will perform 50 random restarts and select the best final alloy composition among them.

Figure 12 shows the results for COBYLA and TRUST-CONSTR. One can observe that the best final alloy compositions obtained by the two algorithms over 50 trials are somewhat different, except that both assign high percentages to Ti and W. The objective function value at the best final alloy composition found by COBYLA is 0.000175099, whereas for TRUST-CONSTR, it is 0.000187352, indicating that the solutions produced by the two algorithms are comparable in terms of cost.

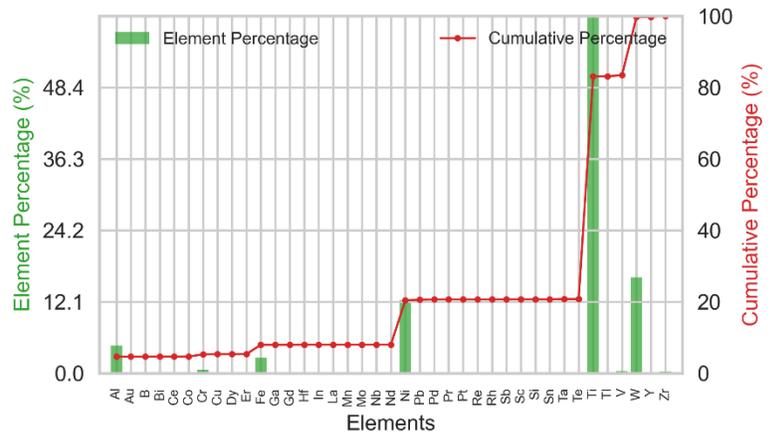
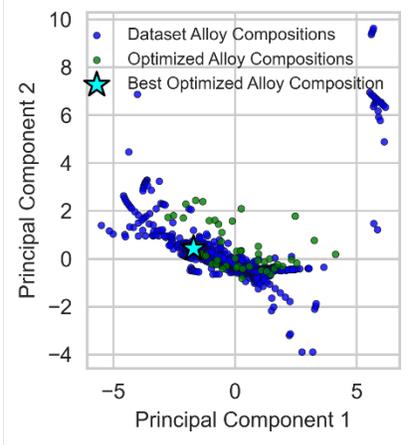
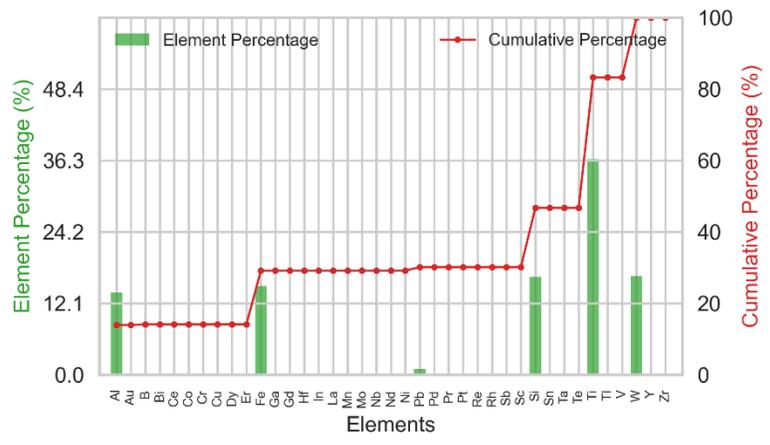
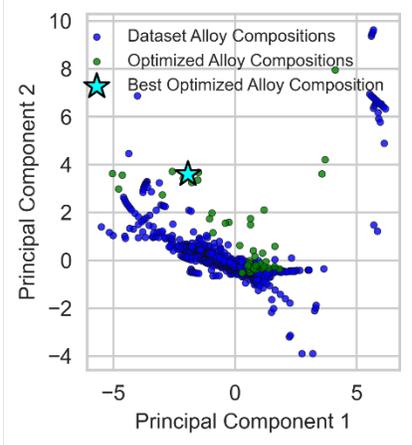

*Figure 12. On the left: the best final alloy compositions obtained by COBYLA (top row) and TRUST-CONSTR (bottom row) over 50 trials. On the right: visualization of the optimization processes.*

*3.4 Achieving desired temperature with minimal alloy cost*

We will now focus on minimizing both objective functions by conducting three experiments: (i) first with $\lambda_1 = 0.25$ and $\lambda_2 = 0.75$, (ii) then with $\lambda_1 = 0.50$ and $\lambda_2 = 0.50$, and (iii) finally with $\lambda_1 = 0.75$ and $\lambda_2 = 0.25$. For each algorithm, we will again perform 50 random restarts and select the best final alloy composition. In the first objective function, we set $T_s = 100$.

Figure 13-Figure 15 present the results for COBYLA and TRUST-CONSTR across the three cases. When $\lambda_1 = 0.25$ and $\lambda_2 = 0.75$, meaning that minimizing cost is prioritized over achieving the desired martensite temperature due to the higher weight of the second objective function, we observe that the best final alloy compositions identified by both algorithms exhibit high levels of Ti and Ni. Ti is the element with the highest percentage, except when $\lambda_1 = 0.75$ and $\lambda_2 = 0.25$. In this case, the percentage of Ti is lower than that of Co in the best final alloy found by COBYLA or comparable to Ni in the best final alloy found by TRUST-CONSTR.

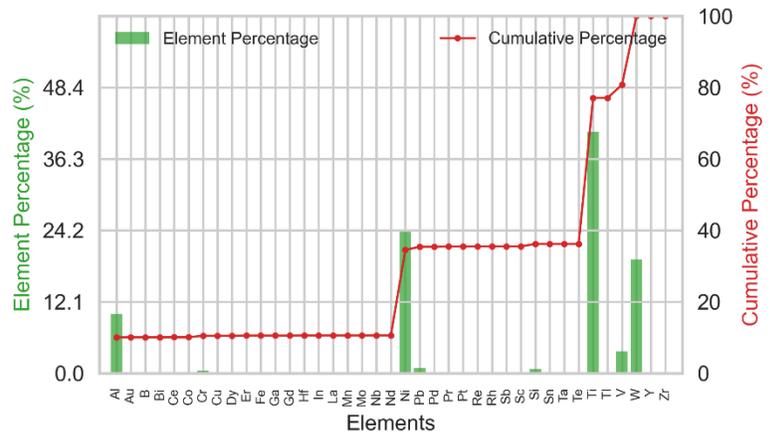
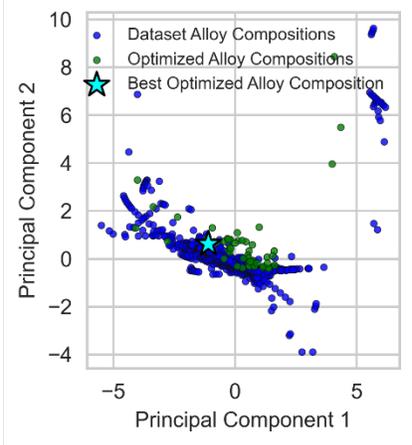
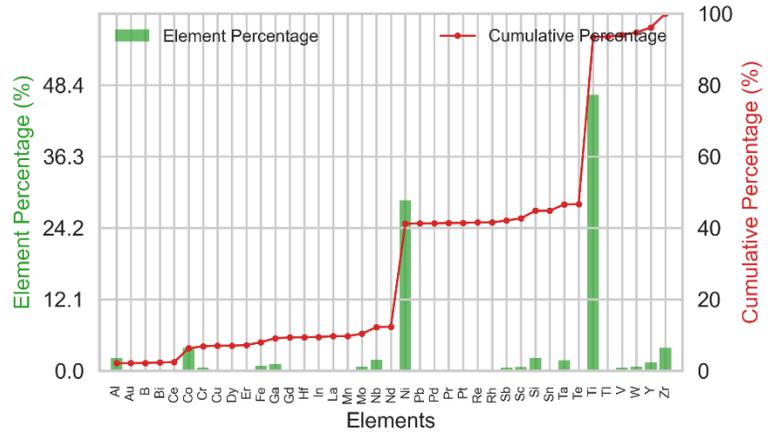
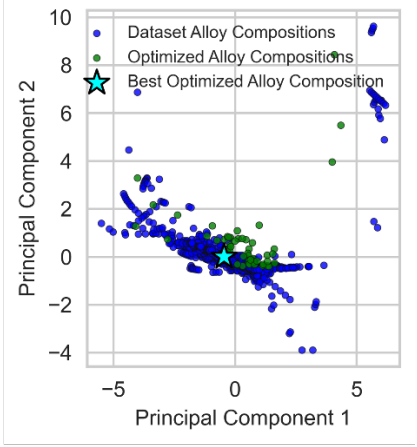

*Figure 13. On the left: the best final alloy compositions obtained by COBYLA (top row) and TRUST-CONSTR (bottom row) over 50 trials with weights $\lambda_1 = 0.25$ and $\lambda_2 = 0.75$. On the right: visualization of the optimization processes.*

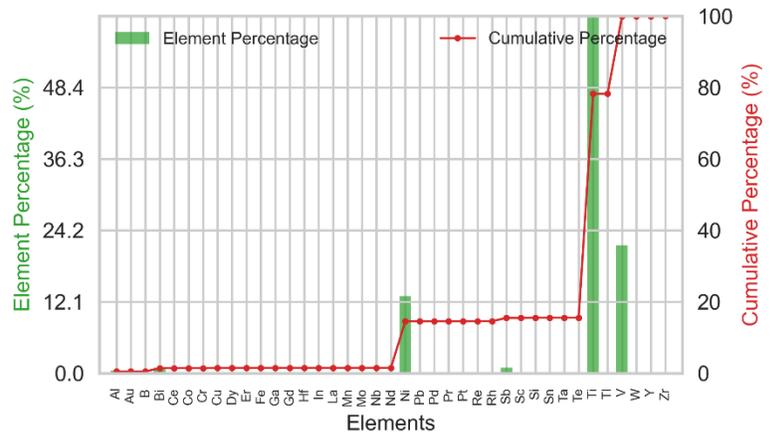
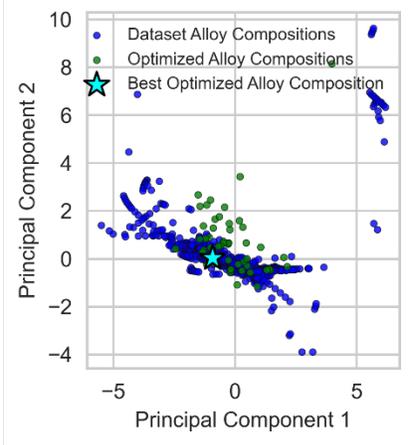
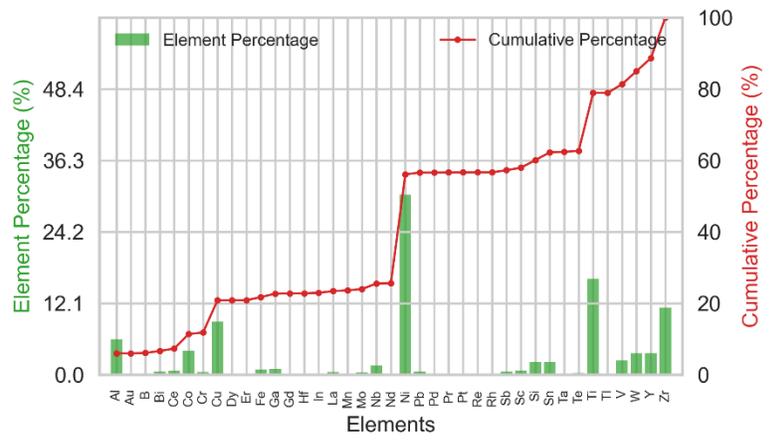
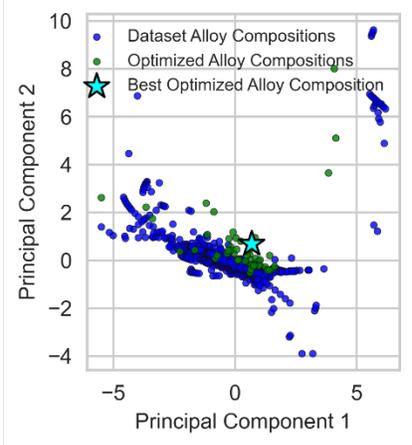

*Figure 14. On the left: the best final alloy compositions obtained by COBYLA (top row) and TRUST-CONSTR (bottom row) over 50 trials with weights $\lambda_1 = 0.5$ and $\lambda_2 = 0.5$. On the right: visualization of the optimization processes.*

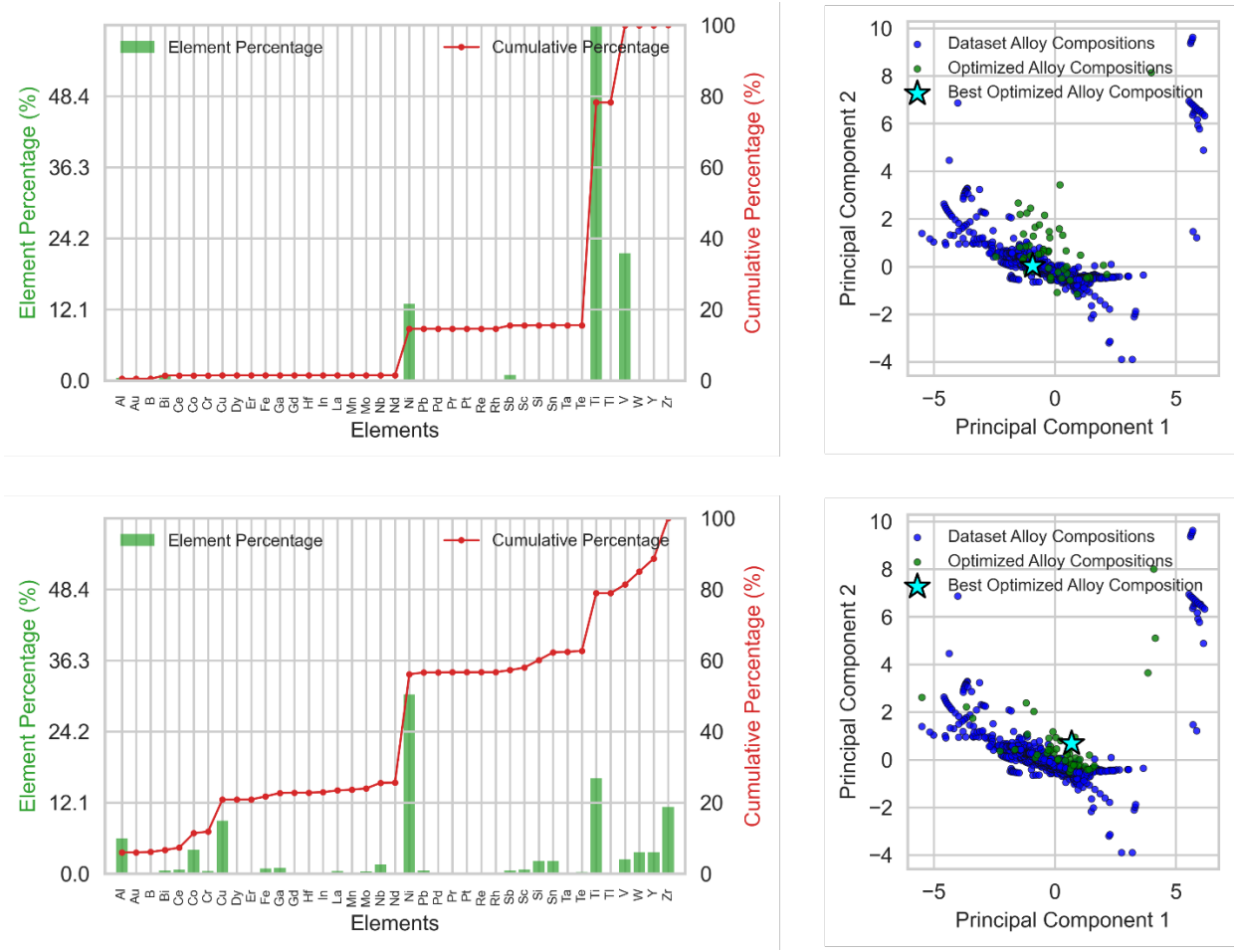

*Figure 15. On the left: the best final alloy compositions obtained by COBYLA (top row) and TRUST-CONSTR (bottom row) over 50 trials with weights $\lambda_1 = 0.75$ and $\lambda_2 = 0.25$. On the right: visualization of the optimization processes.*

## 4. Discussion

This study applied machine learning-based surrogate modeling to explore shape memory alloy (SMA) compositions that balance martensitic start temperature (Ms) performance with cost considerations. As described, we used Extra Trees and neural network models to approximate Ms values and paired these with constrained optimization methods to search for suitable alloy candidates. In this section, we reflect on the modeling choices by comparing our approach with the limited available literature, particularly the multi-objective optimization strategy introduced by Alvi et al. [27].

Alvi et al. highlight a key limitation in standard Bayesian optimization methods, namely that they often model each property independently, even when those properties are related. Their use of Deep Gaussian Processes (DGP-BO) and Multi-Task Gaussian Processes (MTGP-BO) addresses this by jointly modeling multiple outputs, allowing the optimization to take advantage of correlations between properties. In our study, the two objectives Ms and cost are combined using a weighted sum. This method simplifies the problem but does not account for possible relationships between the two objectives. While our approach is suitable for the scale and type of data we used, the MTGP-BO approach may be better suited for more complex design tasks where trade-offs between properties are not independent. Their study also benefits from a significantly larger dataset which was generated through simulations, enabling broader sampling of the compositional space. Furthermore, it should be noted that our method for estimating relative alloy cost only considers the cost of pure elements. The actual production costs are affected by melting method, required cleanliness, thermomechanical processing steps, and finishing operations.

Another aspect of Alvi et al.'s work is their use of a selective sampling strategy (hDGP-BO), which prioritizes querying lower-cost properties to indirectly improve the prediction of more expensive ones. This is especially helpful when data collection is expensive or limited. Our method currently evaluates both objectives for all candidates, which is reasonable given the dataset we used, but more flexible acquisition strategies like theirs could help make the optimization process more efficient in future studies, especially as the number of objectives or data limitations increase.

The use of neural networks in our work builds on existing applications of surrogate modeling in engineering. Previous studies have used neural networks to speed up simulations or predict outcomes in fields like structural mechanics, metal processing, and composite materials. As noted in the earlier sections, these models are most often used for forward prediction, rather than inverse design. While our application to alloy discovery is a step toward filling that gap, the models we used do not account for uncertainty, which makes it harder to guide the optimization process effectively, especially in regions where data is sparse. In contrast, Gaussian process-based models like those used by Alvi et al. include built-in uncertainty estimates that can guide sampling and improve robustness.

Our use of physics-informed features, discussed earlier, reflects a broader trend in materials informatics to combine domain knowledge with machine learning to improve model performance and interpretability. While Alvi et al. do not explicitly include these types of inputs, their modeling framework captures physical relationships implicitly through multi-task learning. These two approaches are complementary: one emphasizes feature construction based on scientific principles, while the other focuses on learning relationships from the data.

## 5. Conclusions

We addressed the challenge of designing high-entropy shape memory alloys (HESMAs) by proposing a surrogate-based optimization framework. The problem was formulated to identify alloy compositions that achieve a target martensitic start temperature while minimizing cost.

- Two surrogate models were used: a tree-based ensemble model paired with the derivative-free COBYLA optimizer, and a neural network model paired with the derivative-based TRUST-CONSTR algorithm.

- While the neural network model showed slightly lower predictive accuracy, its differentiability enabled more reliable and efficient convergence during multi-objective optimization.

- The gradient-based approach consistently identified promising alloy candidates within constrained regions of the compositional space.

- These results highlight the trade-offs between model interpretability, differentiability, and optimization performance in surrogate-based alloy design.

- The framework offers a practical foundation for further development using uncertainty-aware or multi-task surrogate models to address more complex materials discovery challenges.

**Acknowledgements**

The authors would like to express their gratitude to O. Benefan for sharing the NASA dataset with us (for more information, visit https://shapememory.grc.nasa.gov/shape_memory_alloys). Y. Noiman and E. Payton gratefully acknowledge support from NSF Research Experiences for Undergraduate award number EEC-2349580 (REALIZE 2050).**Data Availability**

Page 34 of 39

**Declaration of Competing Interests**

[19] J. Nocedal and S. J. Wright, *Numerical Optimization*. in Springer Series in Operations Research and Financial Engineering. Springer New York, 2006. doi: 10.1007/978-0-387-40065-5.

[20] M. J. D. Powell, "A Direct Search Optimization Method That Models the Objective and Constraint Functions by Linear Interpolation," in *Advances in Optimization and Numerical Analysis*, S. Gomez and J.-P. Hennart, Eds., Dordrecht: Springer Netherlands, 1994, pp. 51–67. doi: 10.1007/978-94-015-8330-5_4.

[21] A. R. Conn, N. I. M. Gould, and P. L. Toint, *Trust Region Methods*. SIAM, 2000.

[22] T. Giovannelli and L. N. Vicente, "An integrated assignment, routing, and speed model for roadway mobility and transportation with environmental, efficiency, and service goals," *Transportation Research Part C: Emerging Technologies*, vol. 152, p. 104144, July 2023, doi: 10.1016/j.trc.2023.104144.

[23] S. J. Honrao, O. Benafan, and J. W. Lawson, "Data-driven study of shape memory behavior of multi-component Ni–Ti alloys in large compositional and processing space," *Shape Memory and Superelasticity*, vol. 9, no. 1, pp. 144–155, 2023.

[24] L. Thiercelin, L. Peltier, and F. Meraghni, "Physics-informed machine learning prediction of the martensitic transformation temperature for the design of 'NiTi-like' high entropy shape memory alloys," *Computational Materials Science*, vol. 231, p. 112578, 2024.

[25] I. T. Jolliffe and J. Cadima, "Principal component analysis: A review and recent developments," *Phil. Trans. R. Soc. A.*, vol. 374, no. 20150202, 2016, doi: https://doi.org/10.1098/rsta.2015.0202.